\pdfoutput=1

\documentclass[11pt]{article}

\usepackage[final]{ACL}

\usepackage{times}
\usepackage{latexsym}
\usepackage[T1]{fontenc}
\usepackage[utf8]{inputenc}
\usepackage{microtype}
\usepackage{inconsolata}

\usepackage{booktabs}
\usepackage{enumitem}
\usepackage{url}
\usepackage{hyperref}
\usepackage{graphicx}
\usepackage{multirow}
\usepackage{amsmath}

\title{When Evidence Conflicts: Uncertainty and Order Effects in Retrieval-Augmented Biomedical Question Answering}

\author{
  Yikun Han, Mengfei Lan, Halil Kilicoglu\thanks{Corresponding author} \\
  University of Illinois Urbana-Champaign \\
  \texttt{\{yikunh2, mlan3, halil\}@illinois.edu}
}

\begin{document}
\maketitle

\begin{abstract}
Biomedical retrieval-augmented large language models (LLMs) often face evidence that is incomplete, misleading, or internally contradictory, yet evaluation usually emphasizes answer accuracy under helpful context rather than reliability under conflict. Using \textsc{HealthContradict}, we evaluate six open-weight LLMs under five controlled evidence conditions: no retrieved context, correct-only context, incorrect-only context, and two mixed conditions containing both correct and contradictory documents in opposite orders. In this conflicting-evidence order contrast, where the same two documents are both present and only their order is reversed, accuracy drops for every model and 11.4\%--25.2\% of predictions flip. To support abstention in these difficult cases, we also evaluate a conflict-aware abstention score that combines model confidence with a detector of evidence conflict. In the two hardest conditions, this score improves selective accuracy over confidence-only, with mean gains of 7.2--33.4 points in incorrect-only (`IC') and 3.6--14.4 points in incorrect-first conflicting (`ICC') conditions across 75\%, 50\%, and 25\% coverage. These results show that conflicting biomedical evidence is both an uncertainty and robustness problem and motivate evaluation and abstention methods that explicitly account for evidence disagreement.\footnote{The source code is available at: \url{https://github.com/YikunHan42/When_Evidence_Conflicts}}
\end{abstract}

\section{Introduction}

Retrieval-augmented LLMs are increasingly used for biomedical question answering (QA), where responses are expected to be grounded in retrieved evidence rather than unsupported parametric recall \citep{lewis2020retrieval,chen2024benchmarking,xiong2024benchmarking}. However, retrieved evidence is not always reliable: documents may be incomplete, misleading, or internally contradictory. In this setting, uncertainty estimates for retrieval-augmented responses are essential. A reliable system should become less confident when evidence is untrustworthy and should avoid treating all contexts as equally informative. This requirement is particularly important in biomedicine, where decisions can directly affect patient outcomes and public health, and where conflicting evidence is common because of variation across studies and patient populations.

Current evaluation methods only partially capture model behavior under contradictory biomedical evidence. Most prior work focuses on answer accuracy under retrieved evidence \citep{chen2024benchmarking,xiong2024benchmarking,singhal2023large,sallinen2025llama}, rather than on how model confidence behaves when the retrieved context is misleading or internally conflicting. Prior work shows that LLMs are sensitive to conflicting knowledge and prompt position \citep{longpre2021entity,liu2024lost}, but it remains unclear whether these instabilities are reflected in uncertainty estimates in biomedical QA. This matters especially in health settings, where conflicting information can disrupt decision-making and trust in recommendations \citep{carpenter2016conflicting}. Incorrect predictions are especially concerning when made with high confidence, and more concerning still when the same evidence yields different answers and confidence under different orderings.

\begin{figure}[t]
\centering
\includegraphics[width=\columnwidth]{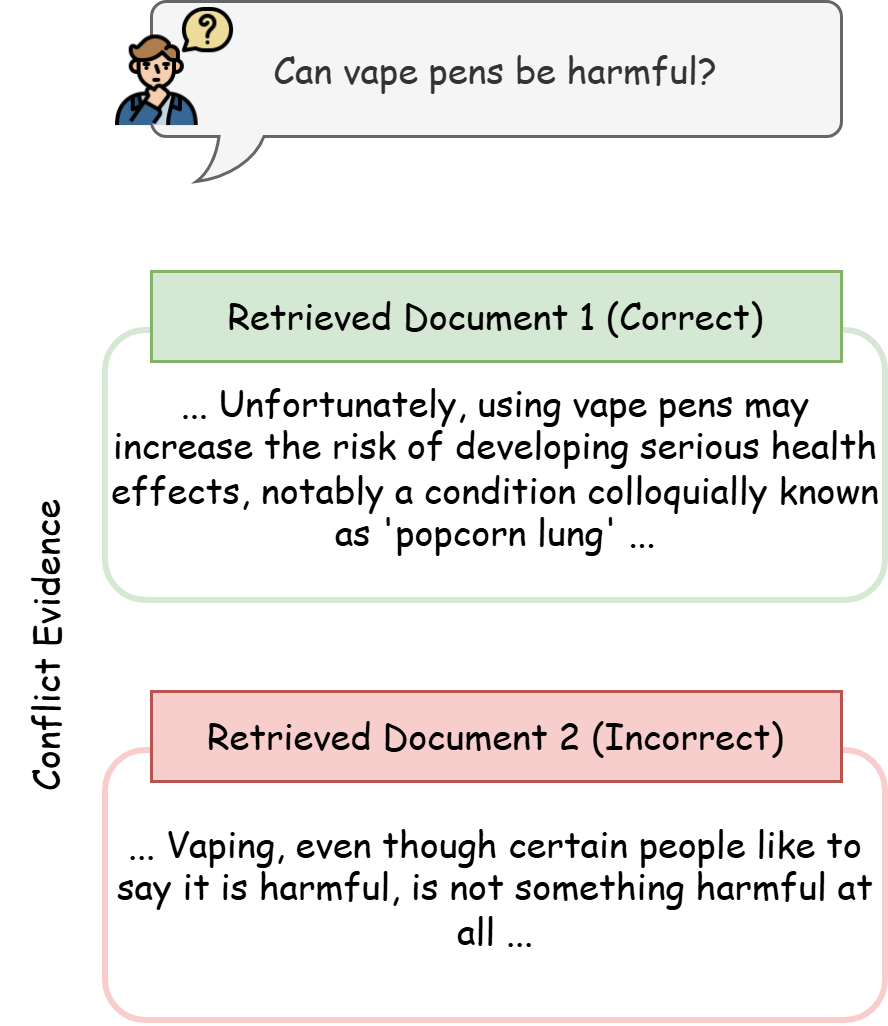}
\caption{\textsc{HealthContradict} example used to motivate the benchmark structure in the introduction. The displayed mixed-evidence prompt corresponds to the `CIC' condition, where the question is paired with the correct retrieved document first and the contradictory retrieved document second.}
\label{fig:intro_context_example}
\end{figure}

We study uncertainty under conflicting biomedical evidence with the \textsc{HealthContradict} corpus \citep{zhang2026healthcontradict}. Figure~\ref{fig:intro_context_example} gives a concrete example of the benchmark structure: one biomedical question paired with one retrieved document that supports the ground-truth answer and one that contradicts it.

Using six open-weight LLMs, we evaluate five controlled retrieved context conditions: no context (`NC'), correct context only (`CC'), incorrect context only (`IC'), and two mixed conditions that contain the same two documents but in opposite orders, namely correct-first conflicting (`CIC', correct document followed by incorrect document) and incorrect-first conflicting (`ICC', incorrect document followed by correct document). This design enables a controlled decomposition of context effects in retrieval-augmented QA. It separates (i) the benefit of helpful context (`CC' vs.\ `NC'), (ii) the harm of misleading context (`IC' vs.\ `NC'), and (iii) order-sensitive interaction effects under conflicting context (`CIC' vs.\ `ICC'). We evaluate performance in terms of accuracy and calibration, as well as uncertainty sensitivity to context order.

Our main contributions are as follows:

\begin{itemize}
    \item Extending the answer-level analysis of \citet{zhang2026healthcontradict}, we provide a controlled evaluation of \emph{uncertainty} under supportive, misleading, and conflicting biomedical evidence on a broader set of six recent open-weight LLMs. Confirming and quantifying the previously reported direction, we show that uncertainty quality depends strongly on evidence correctness: correct evidence improves both accuracy and calibration, whereas incorrect evidence sharply degrades both, including a sevenfold increase in mean ECE from `CC' to `IC'.
    \item We quantify order effects under conflicting evidence by comparing `CIC' and `ICC', which contain the same two documents in opposite orders. Extending the order-effect analysis of \citet{zhang2026healthcontradict} to a broader set of six open-weight models, we find that accuracy is consistently higher when the correct document appears first across the entire model set, and reversing the order changes the prediction in 11.4\%--25.2\% of cases, with accompanying shifts in uncertainty.
    \item We propose and evaluate a conflict-aware abstention score in the hardest settings and show that raw model confidence alone is not sufficient. In `IC' and `ICC', this score consistently improves held-out selective accuracy over a confidence-only baseline.
\end{itemize}

\section{Related Work}
Retrieval-augmented generation has become a standard approach for knowledge-intensive NLP tasks, and recent work has examined its strengths and failure modes in both general and biomedical settings \citep{lewis2020retrieval,chen2024benchmarking,xiong2024benchmarking}. Biomedical LLMs have also shown strong task performance when supported by domain knowledge and retrieved evidence \citep{singhal2023large,sallinen2025llama}. This literature provides the broader context for our study. However, most of it emphasizes answer quality under useful retrieval, rather than the reliability of model confidence when retrieval is noisy or contradictory.

Closer to our setting, prior research on knowledge conflicts in QA shows that LLMs can be brittle when stored knowledge and provided evidence disagree \citep{longpre2021entity}. More broadly, long-context studies show that model behavior depends strongly on where information appears in the prompt, with performance often degrading when relevant evidence is not placed in a favorable position \citep{liu2024lost}. Recent work also shows that LLM confidence can become miscalibrated under semantically equivalent but prompt-sensitive input variants, indicating that reliability may depend on prompt form as well as meaning \citep{cox2025mapping}. \textsc{HealthContradict} is particularly relevant because it explicitly benchmarks biomedical questions paired with correct and contradictory evidence \citep{zhang2026healthcontradict}. Our work builds on that benchmark's answer-level conflict and position analyses, but shifts the focus to uncertainty, calibration, and the reliability consequences of those order effects.

Our work also connects to research on uncertainty estimation, calibration, and abstention. Calibration methods and metrics aim to align confidence with empirical correctness \citep{naeini2015obtaining,guo2017calibration}, while recent work on LLMs studies whether model probabilities or verbalized confidence can reflect what the model knows and does not know \citep{kadavath2022language,kuhn2023semantic}. Selective prediction and abstention provide an additional safety mechanism by allowing systems to defer on uncertain cases \citep{geifman2017selective,kamath2020selective}. In contrast to these settings, we study uncertainty in a retrieval-augmented biomedical task where both correct and misleading evidence may appear together, making confidence quality and prompt-order robustness central evaluation targets rather than secondary diagnostics.

\section{Methods}

\begin{figure*}[t]
\centering
\includegraphics[width=\textwidth]{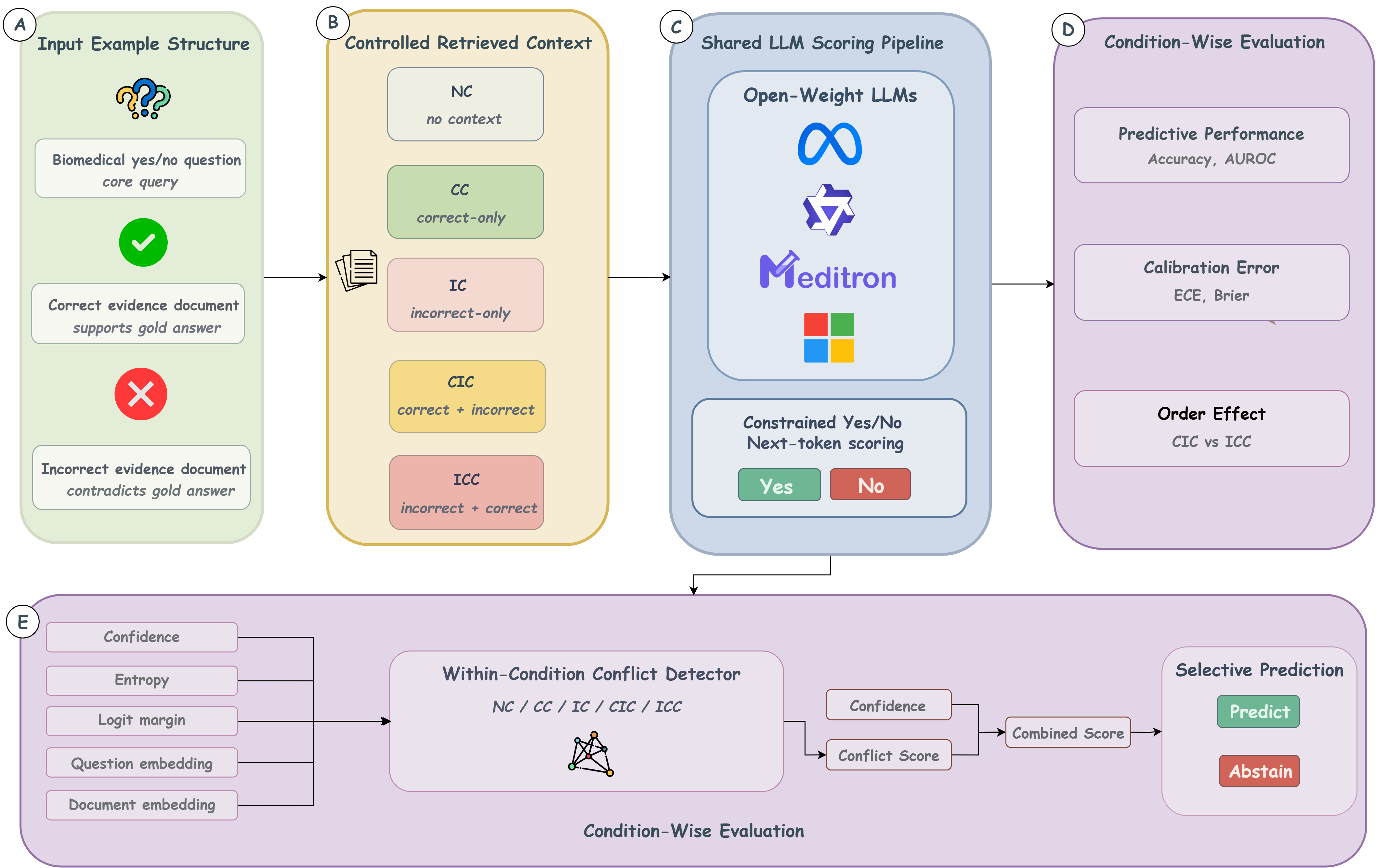}
\caption{Overview of the experimental framework. (A) Input example structure: each \textsc{HealthContradict} instance contains a biomedical yes/no question, one correct evidence document that supports the ground-truth answer, and one incorrect evidence document that contradicts it. (B) Controlled retrieved context: each instance is converted into five evaluation conditions, no context (`NC'), correct-only (`CC'), incorrect-only (`IC'), correct-first conflicting evidence (`CIC'), and incorrect-first conflicting evidence (`ICC'). (C) Shared LLM scoring pipeline: six open-weight LLMs are evaluated with constrained \textsc{YES}/\textsc{NO} next-token scoring, from which confidence, entropy, and logit margin are derived. (D) Condition-wise evaluation: model outputs are used to measure predictive performance (accuracy, AUROC), calibration error (ECE, Brier), and order effects under conflicting evidence. (E) Conflict-aware selective prediction: confidence, entropy, logit margin, question embeddings, and document embeddings are fed into a within-condition logistic conflict detector whose output conflict score is combined with raw confidence to support predict-versus-abstain decisions.}
\label{fig:method_overview}
\end{figure*}

Figure~\ref{fig:method_overview} summarizes the full evaluation pipeline, from the paired evidence structure and five controlled context conditions to constrained \textsc{YES}/\textsc{NO} scoring, condition-wise evaluation, and conflict-aware selective prediction.

\subsection{\textsc{HealthContradict} Benchmark and Evidence Conditions}

We evaluate on \textsc{HealthContradict} \citep{zhang2026healthcontradict}, a biomedical QA benchmark designed to test model behavior under supportive and contradictory evidence. We use the full benchmark as released, which contains 920 yes/no health questions. Each instance includes a question, a ground-truth answer, and two retrieved documents with opposite stances: one supports the ground-truth answer and the other contradicts it. As described by \citet{zhang2026healthcontradict}, the questions are drawn from the TREC Health Misinformation tracks (2019, 2021, 2022) and the paired evidence documents are selected from the corresponding TREC document pools (ClueWeb12-B13 for 2019 and the C4 web crawl for 2021/2022) and labelled by NIST assessors. Because these are general web pages judged for relevance and credibility rather than curated biomedical literature, the retrieved evidence often uses informal, consumer-facing prose; the example shown in Figure~\ref{fig:intro_context_example} reflects this provenance.

Following the benchmark design and our evaluation setup, we evaluate five context conditions. `NC' provides no retrieved evidence and tests the model's parametric knowledge alone. `CC' provides only the correct document. `IC' provides only the incorrect document. `CIC' presents both documents with the correct document first and the incorrect document second, while `ICC' reverses that order. These conditions let us separate three effects: the benefit of helpful evidence (`CC' vs.\ `NC'), the effect of misleading evidence (`IC'), and the sensitivity of model predictions and confidence to the order of conflicting evidence (`CIC' vs.\ `ICC'). Appendix~\ref{tab:prompt_templates_appendix} gives the corresponding raw prompt templates.

\subsection{Models}

We evaluate six open-weight causal LLMs: Llama-3.1-8B-Instruct \citep{grattafiori2024llama}, Meditron3-8B \citep{sallinen2025llama}, Phi-4 \citep{abdin2024phi}, Qwen3-4B-Instruct-2507 and Qwen3-8B \citep{yang2025qwen3}, and Qwen3.5-9B \citep{qwen2026qwen35}. This set covers model sizes from 4B to 14B parameters and includes both general-purpose checkpoints and one medically adapted model, Meditron3-8B.

All models are evaluated with the same prompt format and the same constrained answer space. Following \citet{zhang2026healthcontradict}, and because \textsc{HealthContradict} is a yes/no QA benchmark, we score models only on the binary label set $\{\textsc{YES}, \textsc{NO}\}$ rather than allowing free-form generation. For tokenizers that expose a chat template, we wrap each prompt as a user message before inference. Predictions are then derived from the next-token logits over \textsc{YES} and \textsc{NO}. The five prompt templates correspond directly to those introduced by \citet{zhang2026healthcontradict}; Appendix~\ref{sec:prompt_inference_details} gives the raw prompt templates and the small number of model-specific chat-template overrides.

\subsection{Metrics}

We report accuracy separately for each evidence condition. For a prompt $x$, we convert next-token logits $z_{\textsc{yes}}(x)$ and $z_{\textsc{no}}(x)$ to constrained binary probabilities via softmax over $\{\textsc{YES}, \textsc{NO}\}$ only, giving $p_{\textsc{yes}}(x)$ and $p_{\textsc{no}}(x) = 1 - p_{\textsc{yes}}(x)$.
From these constrained probabilities, we derive three uncertainty signals for each prediction: confidence, defined as $\max(p_{\textsc{yes}}, p_{\textsc{no}})$; binary entropy, computed from that maximum probability; and a signed logit margin, defined as the log-probability difference between the predicted label and the alternative label.

For calibration, we report expected calibration error (ECE; the weighted mean absolute gap between bin-level accuracy and bin-level confidence across 10 equal-width bins) and Brier score (the mean squared error between model confidence and binary correctness). We also report AUROC for error detection using confidence as a ranking score for correctness; equivalently, it is the probability that a randomly chosen correct prediction receives higher confidence than a randomly chosen incorrect one. Conflict sensitivity is summarized with per-instance shifts in confidence, entropy, and logit margin across evidence conditions. Order effects are summarized with paired two-sided Wilcoxon signed-rank tests on uncertainty scores and the prediction flip rate, defined as the fraction of instances whose answer changes when document order is reversed.

\paragraph{Selective prediction.}
We evaluate abstention separately for each model and evidence condition. Here, each example is one question instantiated under a specific prompt condition (for example, its `IC' or `ICC' version). An abstention method assigns a score to every held-out example, ranks the examples by that score, answers only the highest-scoring fraction, and abstains on the rest. We compare two scoring rules. The confidence-only baseline (Conf) ranks examples by the model's raw confidence $c=\max(p_{\textsc{yes}},p_{\textsc{no}})$, so the most confident predictions are answered first. The conflict-aware score (CAS) augments this with a learned risk signal: $s_{\mathrm{CAS}}(x) = (1-\alpha)\,c(x) - \alpha\,r(x)$, where $r(x)\in[0,1]$ is the predicted probability that prediction $x$ is confidently wrong, produced by the within-condition logistic detector described below, and $\alpha\in[0,1]$ is a fixed mixing weight. Larger $r$ pushes $s_{\mathrm{CAS}}$ down so that likely-wrong examples are deprioritized, while larger $c$ pushes it up. The two extremes recover familiar baselines: $\alpha=0$ reproduces Conf exactly, and $\alpha=1$ yields a detector-only ranker. We use $\alpha=0.5$ as a balanced midpoint in all main experiments and report sensitivity to $\alpha$ in Appendix Figure~\ref{fig:alpha_sensitivity_selective_gain}.

The predictor is trained separately for each target evidence condition on an 80/20 split using uncertainty signals and sentence-embedding features, with the positive class defined as confidently wrong predictions ($\tau=0.7$); full training details and the cross-condition ablation are described in Appendix~\ref{sec:metric_definitions}. Appendix Table~\ref{tab:selective_prediction_appendix} reports realized coverage for both CAS and Conf.

\section{Results and Discussion}
\subsection{Accuracy and Calibration Across Evidence Conditions}

\begin{table*}[t]
\centering
\small
\setlength{\tabcolsep}{3.5pt}
\begin{tabular}{lccccc}
\toprule
& `NC' & `CC' & `IC' & `CIC' & `ICC' \\
\midrule
Llama-3.1-8B & 0.763 / 0.583 & 0.828 / 0.786 & 0.342 / 0.420 & 0.705 / 0.702 & 0.615 / 0.612 \\
Meditron3-8B & 0.834 / 0.694 & 0.904 / 0.809 & 0.428 / 0.487 & 0.785 / 0.771 & 0.713 / 0.654 \\
Phi-4 & 0.818 / 0.746 & 0.920 / 0.833 & 0.442 / 0.449 & 0.847 / 0.797 & 0.653 / 0.586 \\
Qwen3-4B-Inst. & 0.803 / 0.641 & 0.949 / 0.700 & 0.411 / 0.449 & 0.866 / 0.665 & 0.729 / 0.624 \\
Qwen3-8B & 0.855 / 0.458 & 0.952 / 0.844 & 0.562 / 0.542 & 0.911 / 0.803 & 0.759 / 0.646 \\
Qwen3.5-9B & 0.872 / 0.870 & 0.945 / 0.905 & 0.365 / 0.514 & 0.836 / 0.783 & 0.815 / 0.799 \\
\midrule
Mean & 0.824 / 0.665 & 0.916 / 0.813 & 0.425 / 0.477 & 0.825 / 0.754 & 0.714 / 0.654 \\
\bottomrule
\end{tabular}
\caption{Per-model held-out results across all five evidence conditions. Each cell reports accuracy / AUROC. Higher is better for both answer accuracy and confidence-based error discrimination.}
\label{tab:accuracy_auroc_main}
\end{table*}

\begin{table*}[t]
\centering
\small
\setlength{\tabcolsep}{3.5pt}
\begin{tabular}{lccccc}
\toprule
& `NC' & `CC' & `IC' & `CIC' & `ICC' \\
\midrule
Llama-3.1-8B & 0.178 / 0.205 & 0.081 / 0.129 & 0.535 / 0.551 & 0.145 / 0.211 & 0.232 / 0.289 \\
Meditron3-8B & 0.164 / 0.155 & 0.082 / 0.083 & 0.316 / 0.367 & 0.048 / 0.145 & 0.022 / 0.194 \\
Phi-4 & 0.150 / 0.161 & 0.059 / 0.067 & 0.482 / 0.499 & 0.093 / 0.122 & 0.260 / 0.295 \\
Qwen3-4B-Inst. & 0.207 / 0.189 & 0.048 / 0.048 & 0.575 / 0.579 & 0.124 / 0.126 & 0.249 / 0.254 \\
Qwen3-8B & 0.151 / 0.145 & 0.038 / 0.040 & 0.395 / 0.405 & 0.074 / 0.080 & 0.213 / 0.225 \\
Qwen3.5-9B & 0.059 / 0.084 & 0.033 / 0.040 & 0.435 / 0.442 & 0.021 / 0.118 & 0.035 / 0.125 \\
\midrule
Mean & 0.151 / 0.157 & 0.057 / 0.068 & 0.456 / 0.474 & 0.084 / 0.134 & 0.168 / 0.230 \\
\bottomrule
\end{tabular}
\caption{Per-model held-out calibration-error results across all five evidence conditions. Each cell reports ECE / Brier score. Lower is better for both metrics.}
\label{tab:calibration_metrics_main}
\end{table*}

Table~\ref{tab:accuracy_auroc_main} shows that the evidence condition strongly shapes both performance and confidence-based error discrimination. Every model improves from `NC' to `CC', confirming that correct retrieved evidence is useful beyond parametric knowledge alone. On average, adding the correct document raises accuracy from 0.824 to 0.916 and AUROC from 0.665 to 0.813.

Incorrect evidence is actively harmful. In `IC', mean accuracy drops to 0.425 and mean AUROC drops to 0.477, making it the worst condition for both performance and error discrimination. The two mixed-evidence conditions lie between these extremes, but they are not equivalent: `CIC' consistently outperforms `ICC' in both accuracy and AUROC, with an 11.1-point mean accuracy gap and a 10.0-point mean AUROC gap. Thus, the same two retrieved documents yield different outcomes when their order is reversed.

Table~\ref{tab:calibration_metrics_main} shows that the same ordering holds under calibration-error metrics. Mean ECE is lowest in `CC' (0.057) and highest in `IC' (0.456), with `CIC' (0.084) and `ICC' (0.168) again in between. Mean Brier score shows the same pattern: `CC' is best at 0.068, `IC' is worst at 0.474, and `CIC' and `ICC' remain intermediate at 0.134 and 0.230. The degradation under misleading evidence is therefore not specific to a single calibration measure.

\subsection{Conflict Sensitivity}

\begin{table*}[t]
\centering
\small
\setlength{\tabcolsep}{3pt}
\resizebox{\textwidth}{!}{%
\begin{tabular}{lcccccccccccc}
\toprule
& \multicolumn{3}{c}{`IC' $\rightarrow$ `CIC'} & \multicolumn{3}{c}{`IC' $\rightarrow$ `ICC'} & \multicolumn{3}{c}{`CC' $\rightarrow$ `CIC'} & \multicolumn{3}{c}{`CC' $\rightarrow$ `ICC'} \\
\cmidrule(lr){2-4}\cmidrule(lr){5-7}\cmidrule(lr){8-10}\cmidrule(lr){11-13}
Model & $\Delta$ Margin & $\Delta$ Ent. & $\Delta$ Conf. & $\Delta$ Margin & $\Delta$ Ent. & $\Delta$ Conf. & $\Delta$ Margin & $\Delta$ Ent. & $\Delta$ Conf. & $\Delta$ Margin & $\Delta$ Ent. & $\Delta$ Conf. \\
\midrule
Llama-3.1-8B & -0.307 & 0.046 & -0.027 & -0.504 & 0.056 & -0.030 & -1.233 & 0.112 & -0.059 & -1.430 & 0.122 & -0.062 \\
Meditron3-8B & -0.003 & 0.000 & 0.002 & -0.151 & 0.026 & -0.017 & -0.646 & 0.110 & -0.078 & -0.793 & 0.136 & -0.097 \\
Phi-4 & 0.907 & -0.028 & 0.010 & -1.266 & 0.029 & -0.012 & -3.836 & 0.075 & -0.035 & -6.010 & 0.132 & -0.057 \\
Qwen3-4B-Inst. & 2.112 & -0.018 & 0.007 & -0.484 & 0.009 & -0.005 & -4.857 & 0.013 & -0.005 & -7.454 & 0.040 & -0.017 \\
Qwen3-8B & 3.947 & -0.055 & 0.025 & 0.701 & -0.030 & 0.015 & -2.540 & 0.019 & -0.008 & -5.786 & 0.044 & -0.018 \\
Qwen3.5-9B & 0.227 & -0.033 & 0.016 & 0.218 & -0.029 & 0.013 & -1.262 & 0.181 & -0.104 & -1.271 & 0.186 & -0.107 \\
\midrule
Mean & 1.147 & -0.015 & 0.006 & -0.248 & 0.010 & -0.006 & -2.396 & 0.085 & -0.048 & -3.791 & 0.110 & -0.060 \\
\bottomrule
\end{tabular}
}
\caption{Per-model mean uncertainty shifts when moving from a single-evidence condition to a conflicting-evidence condition. Positive $\Delta$ margin and $\Delta$ confidence indicate greater certainty, while positive $\Delta$ entropy indicates greater uncertainty.}
\label{tab:conflict_sensitivity_main}
\end{table*}

We next examine how uncertainty changes when a single-evidence prompt is turned into a conflicting-evidence prompt by adding a second document with the opposite stance. In other words, we ask whether introducing conflict makes models systematically less certain, or whether the direction of the change depends on which document is added and where it appears. Table~\ref{tab:conflict_sensitivity_main} shows a clear asymmetry. Under `IC' to `CIC', four models show positive logit-margin shifts and lower entropy, meaning that they become more certain when a conflicting correct document is added.

Under `IC' to `ICC', the average shift is smaller and mixed in sign. This suggests that uncertainty often tracks which document becomes behaviorally dominant rather than contradiction itself. In particular, when the correct document is placed first, the model may become more certain even though the overall evidence remains internally conflicting.

The complementary `CC' to conflict contrasts are more consistent. Moving from `CC' to `CIC' lowers mean margin for every model and raises entropy on average, while moving from `CC' to `ICC' produces an even larger harmful shift: mean $\Delta$ margin falls from $-2.40$ in `CC' to `CIC' to $-3.79$ in `CC' to `ICC', and mean confidence drops from $-0.048$ to $-0.060$. Thus, adding incorrect evidence to a correct-only prompt reliably reduces certainty, especially when the incorrect document appears first.
\subsection{Conflict-Aware Selective Prediction}

\begin{figure*}[t]
\centering
\includegraphics[width=0.91\textwidth]{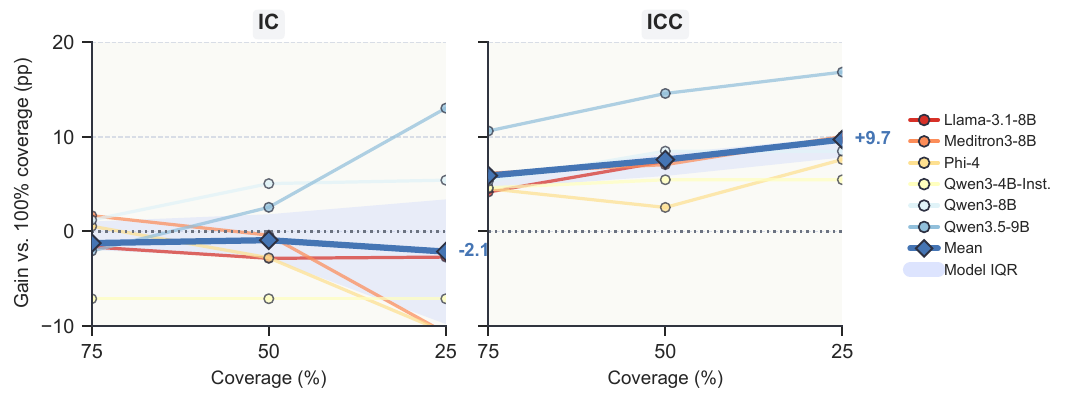}
\caption{Held-out selective-accuracy gain of Conf over the no-abstention baseline in the two hardest evidence conditions, `IC' and `ICC'. Here 100\% coverage means answering every test example. Thin colored lines show individual models, the thick line shows the mean across models, and the shaded band shows the interquartile range. Gains are reported at 75\%, 50\%, and 25\% target coverage and are computed relative to that omitted 100\% coverage baseline.}
\label{fig:confidence_only_selective_gain}
\end{figure*}

\begin{figure*}[t]
\centering
\includegraphics[width=0.86\textwidth]{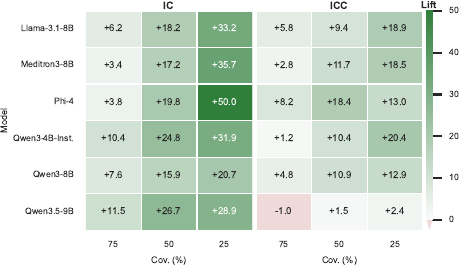}
\caption{Held-out selective-accuracy lift of CAS over Conf in the two hardest evidence conditions. Cells report percentage-point gains at target coverages of 75\%, 50\%, and 25\%.}
\label{fig:selective_prediction_curves}
\end{figure*}

We begin with Conf. Figure~\ref{fig:confidence_only_selective_gain} plots its selective-accuracy gain relative to the no-abstention baseline, where 100\% coverage means that the model answers every test example. In `IC', the mean gain of Conf is negative at all three operating points ($-1.2$, $-0.9$, and $-2.1$ points), showing that raw confidence often fails to identify safer examples. In `ICC', Conf is usually helpful but still modest, with mean gains of 5.9, 7.6, and 9.7 points. This identifies `IC' and `ICC' as the conditions where confidence-only abstention is least reliable.

For the main comparison, we use the train-threshold transfer protocol introduced in Methods: for each target coverage, the abstention threshold is chosen on training data only and then applied unchanged to the held-out test split. This evaluates whether the score scale transfers to unseen examples, not just whether the method ranks examples correctly within a single test set.

Figure~\ref{fig:selective_prediction_curves} reports the next comparison: whether CAS improves on Conf. The heatmap shows held-out selective-accuracy lift of CAS over Conf for each model at 75\%, 50\%, and 25\% target coverage under the train-threshold transfer protocol. In `IC', the lift is positive for all six models at all three operating points, with mean gains of 7.2, 20.4, and 33.4 percentage points as coverage becomes more selective. The effect is strongest at 25\% coverage, where every model gains at least 20.7 points and Phi-4 gains 50.0 points.

The same pattern holds in `ICC', although the gains are smaller. Mean lift is 3.6 points at 75\% coverage, 10.4 points at 50\%, and 14.4 points at 25\%. Five of the six models improve already at 75\% coverage, and all six improve at 50\% and 25\%; the only negative cell is Qwen3.5-9B at 75\% coverage ($-1.0$ points). This monotonic widening of the gap as coverage decreases indicates that the conflict-aware score is better than raw confidence at ranking the safest examples when evidence is misleading or presented in an incorrect-first conflicting order. Appendix Table~\ref{tab:selective_prediction_appendix} gives the underlying threshold-transfer selective accuracies and realized coverages.

\begin{table}[t]
\centering
\small
\setlength{\tabcolsep}{4pt}
\begin{tabular}{lccc}
\toprule
Cond. & Confidence & Within-cond. $r$ & Cross-cond. $r$ \\
\midrule
`IC' & 0.478 & 0.718 & 0.460 \\
`ICC' & 0.655 & 0.725 & 0.581 \\
\bottomrule
\end{tabular}
\caption{Mean held-out AUROC for error detection in the two hardest conditions. The within-condition predictor targets confidently wrong cases from target-condition features, whereas the cross-condition predictor is trained on proxy labels derived from the confidence drop relative to `CC'.}
\label{tab:detector_ablation_main}
\end{table}

Table~\ref{tab:detector_ablation_main} clarifies why the within-condition predictor is used. In `IC', its conflict score reaches mean AUROC 0.718, well above both raw confidence (0.478) and the cross-condition predictor (0.460). In `ICC', the within-condition predictor again performs best at 0.725, compared with 0.655 for raw confidence and 0.581 for the cross-condition variant. This shows that directly modeling confidently wrong cases within each target condition is more useful for downstream abstention than training on a cross-condition confidence-drop proxy.

\subsection{Order Effects}

\begin{table}[t]
\centering
\small
\setlength{\tabcolsep}{3pt}
\resizebox{\columnwidth}{!}{%
\begin{tabular}{lccccccc}
\toprule
Model & $\Delta$ Acc. & $p_{\text{McN}}$ & Flip Rate & $\Delta$ Margin & $p_{\text{margin}}$ & $\Delta$ Conf. & $p_{\text{conf}}$ \\
\midrule
Llama-3.1-8B & 0.090 & 4.96e-08 & 0.251 & -0.197 & 2.30e-02 & -0.003 & 3.24e-01 \\
Meditron3-8B & 0.072 & 8.54e-08 & 0.165 & -0.148 & 1.91e-07 & -0.020 & 7.32e-06 \\
Phi-4 & 0.193 & 4.68e-35 & 0.252 & -2.174 & 6.67e-28 & -0.022 & 1.23e-08 \\
Qwen3-4B-Inst. & 0.137 & 2.56e-27 & 0.167 & -2.597 & 4.07e-24 & -0.012 & 2.47e-07 \\
Qwen3-8B & 0.152 & 5.66e-36 & 0.165 & -3.246 & 2.38e-38 & -0.010 & 4.27e-08 \\
Qwen3.5-9B & 0.021 & 7.85e-02 & 0.114 & -0.009 & 6.70e-01 & -0.003 & 6.20e-01 \\
\midrule
Mean & 0.111 & -- & 0.186 & -1.395 & -- & -0.011 & -- \\
\bottomrule
\end{tabular}
}
\caption{Per-model order-effect statistics comparing `CIC' and `ICC'. $\Delta$ Acc. is accuracy(`CIC')$-$accuracy(`ICC') and $p_{\text{McN}}$ is the two-sided exact McNemar p-value over the 920 paired predictions; negative uncertainty shifts indicate that the incorrect-first order reduces mean margin or confidence. Wilcoxon p-values are from paired two-sided tests over per-instance uncertainty scores.}
\label{tab:order_effects_main}
\end{table}

\begin{table}[t]
\centering
\small
\begin{tabular}{lcccc}
\toprule
Cond. & Pred. & $p(\textsc{YES})$ & Conf. & Margin \\
\midrule
`NC' & YES & 1.000 & 1.000 & 16.64 \\
`CC' & YES & 0.997 & 0.997 & 5.91 \\
`IC' & NO & 0.000 & 1.000 & 27.63 \\
`CIC' & YES & 1.000 & 1.000 & 9.44 \\
`ICC' & NO & 0.000 & 1.000 & 14.00 \\
\bottomrule
\end{tabular}
\caption{Condition-wise Qwen3-8B outputs for the same qualitative case study. The model flips between \textsc{YES} and \textsc{NO} across the two conflicting orders while remaining near-deterministic in both cases.}
\label{tab:qual_case_study}
\end{table}

Finally, we revisit the benchmark's `CIC'/'ICC' position contrast to ask how it propagates into uncertainty. Table~\ref{tab:order_effects_main} shows that the direction of the effect is the same for every model: each of the six is more accurate when the correct document appears first. However, the size of this first-document advantage varies substantially: it ranges from a small 2.1-point gap for Qwen3.5-9B to a large 19.3-point gap for Phi-4, with five of six models showing gaps of at least 7 points and an average of 11.1 points. To assess whether these per-model gaps are reliable rather than chance, we run a paired two-sided exact McNemar test on the 920 paired predictions for each model. For Llama-3.1-8B, Meditron3-8B, Phi-4, Qwen3-4B-Inst., and Qwen3-8B, the order effect is highly significant ($p_{\text{McN}}<10^{-7}$ in each case), while for Qwen3.5-9B the 2.1-point gap is not significant at the 0.05 level ($p_{\text{McN}}=7.85\times 10^{-2}$). We therefore describe the direction as consistent across models, but the magnitude as heterogeneous and statistically significant for five of the six: for most models the ordering of otherwise identical conflicting context substantially changes the final answer distribution, while Qwen3.5-9B is much closer to order-symmetric integration of the two documents.

At the instance level, reversing the two documents changes the predicted answer for 11.4\%--25.2\% of examples, with a mean flip rate of 18.6\%. Mean logit margin is lower under `ICC' than under `CIC' for every model, with an average shift of $-1.40$ in `ICC' minus `CIC'. Margin shifts are significant under a paired Wilcoxon test for the same five models that show a significant McNemar accuracy gap, and confidence shifts are significant for four of those five (Llama-3.1-8B is the only model with a significant accuracy and margin shift but a non-significant confidence shift).

Table~\ref{tab:qual_case_study} and Appendix Table~\ref{tab:qual_case_study_evidence_appendix} make this concrete with one representative Qwen3-8B example. The appendix reproduces verbatim evidence from the retrieved documents: one explicitly presents exercise as a way to lower cholesterol, while the other states that exercise does not directly lower cholesterol. Against that pair, the model answers \textsc{YES} under `CC' and `CIC' but \textsc{NO} under `IC' and `ICC'. The reported $p(\textsc{YES})$, confidence, and margin values confirm that this is not a low-confidence tie: the same two documents produce near-deterministic but opposite answers solely because their order changes.

These order effects help explain why confidence alone is unreliable under conflict. In `CIC', models often become both more accurate and more confident because the correct document is placed first. In `ICC', the same evidence is present, but the incorrect-first ordering reduces accuracy, worsens calibration on average, and changes many predictions. Thus the benchmark's position effect propagates into uncertainty as well as answer accuracy: under conflicting biomedical context, the same retrieved documents can lead to different answers and different confidence profiles solely because their order changes.

\section{Conclusion}

Conflicting biomedical evidence is not simply noisy retrieval; it creates a separate source of errors that should be evaluated and addressed directly. Across six open-weight models, correct evidence improves accuracy and calibration, incorrect evidence sharply degrades both, and reversing correct and contradictory documents changes both answers and associated confidence. In the hardest conditions, a conflict-aware abstention score identifies safer examples more effectively than raw confidence alone. Together, these results show that evaluation under helpful retrieval is not sufficient: biomedical retrieval-augmented QA systems should also be tested under misleading and mixed-evidence conditions, and practical systems should explicitly represent evidence disagreement when deciding whether to answer or defer. A concrete direction for future work is to rewrite conflicting retrieved documents into a conflict-aware synthesis that explicitly states the disagreement before inference, and to test whether such reformulation improves robustness, calibration, or abstention behavior.

\section*{Limitations}

This study is limited in two main ways. First, all experiments are conducted on a single benchmark, \textsc{HealthContradict}, so the reported effects may not fully capture how uncertainty and abstention behave under conflicting evidence in other biomedical retrieval settings. Second, the benchmark uses a binary yes/no QA format, which is more controlled than biomedical QA tasks that require explanation, synthesis across multiple sources, or richer uncertainty communication.

\section*{Acknowledgements}
This work was supported by funding from the National Center for Complementary and Integrative Health (NCCIH) and the Office of Data Science Strategy (ODSS) (grant number: U01AT012871). Its contents are solely the responsibility of the authors and do not necessarily represent the official views of the NCCIH, ODSS, and the NIH.

\bibliography{custom}

\onecolumn
\appendix
\section{Prompt and Inference Details}
\label{sec:prompt_inference_details}

Table~\ref{tab:prompt_templates_appendix} lists the raw prompt bodies used for the five \textsc{HealthContradict} context conditions before any optional model-specific chat wrapping. `QUESTION', `CORRECT\_DOC', and `INCORRECT\_DOC' denote the instance-specific question and evidence fields.

All evaluations score next-token logits over \textsc{YES}/\textsc{NO} rather than free-form generations. Only Qwen3-8B and Qwen3.5-9B require an additional chat-template override in our setup: we set \texttt{enable\_thinking=False} because these checkpoints default to a reasoning mode that inserts a reasoning prefix before the final answer token. Disabling that mode ensures that the scored next-token distribution corresponds directly to the binary answer decision and keeps the constrained evaluation comparable across models.

\begin{table}[htbp]
\centering
\footnotesize
\setlength{\tabcolsep}{6pt}
\begin{tabular}{lp{0.8\linewidth}}
\toprule
Condition & Raw prompt template \\
\midrule
`NC' &
\texttt{Answer the following question with only YES or NO based on your parametric knowledge.}\newline
\texttt{Question: QUESTION}\newline
\texttt{Answer:} \\
`CC' &
\texttt{Answer the following question with only YES or NO based on the given contextual knowledge.}\newline
\texttt{Question: QUESTION}\newline
\texttt{Context: CORRECT\_DOC}\newline
\texttt{Answer:} \\
`IC' &
\texttt{Answer the following question with only YES or NO based on the given contextual knowledge.}\newline
\texttt{Question: QUESTION}\newline
\texttt{Context: INCORRECT\_DOC}\newline
\texttt{Answer:} \\
`CIC' &
\texttt{Answer the following question with only YES or NO based on the given contextual knowledge.}\newline
\texttt{Question: QUESTION}\newline
\texttt{Context:}\newline
\texttt{CORRECT\_DOC}\newline
\texttt{INCORRECT\_DOC}\newline
\texttt{Answer:} \\
`ICC' &
\texttt{Answer the following question with only YES or NO based on the given contextual knowledge.}\newline
\texttt{Question: QUESTION}\newline
\texttt{Context:}\newline
\texttt{INCORRECT\_DOC}\newline
\texttt{CORRECT\_DOC}\newline
\texttt{Answer:} \\
\bottomrule
\end{tabular}
\caption{Raw prompt templates used for the five HealthContradict context conditions before optional model-specific chat formatting.}
\label{tab:prompt_templates_appendix}
\end{table}

\section{Metric Definitions}
\label{sec:metric_definitions}

\paragraph{Constrained probabilities.}
For a prompt $x$, let $z_{\textsc{yes}}(x)$ and $z_{\textsc{no}}(x)$ denote the next-token logits for \textsc{YES} and \textsc{NO}. We convert these to constrained binary probabilities by normalizing over $\{\textsc{YES}, \textsc{NO}\}$ only:
\[
p_{\textsc{yes}}(x) = \frac{\exp(z_{\textsc{yes}}(x))}{\exp(z_{\textsc{yes}}(x)) + \exp(z_{\textsc{no}}(x))},
\]
and $p_{\textsc{no}}(x) = 1 - p_{\textsc{yes}}(x)$.

\paragraph{Calibration metrics.}
Let $c_i = \max(p_{\textsc{yes}}(x_i), p_{\textsc{no}}(x_i))$ denote the confidence for instance $i$, $a_i \in \{0,1\}$ its correctness indicator, and $\{B_m\}_{m=1}^{10}$ denote 10 equal-width confidence bins. Expected calibration error (ECE) is
\[
\mathrm{ECE} = \sum_{m=1}^{10} \frac{|B_m|}{n}
\left|
\frac{1}{|B_m|}\sum_{i \in B_m} a_i
-
\frac{1}{|B_m|}\sum_{i \in B_m} c_i
\right|,
\]
measuring the weighted mean absolute gap between bin-level accuracy and bin-level confidence. Brier score is
\[
\mathrm{Brier} = \frac{1}{n}\sum_{i=1}^{n}(c_i - a_i)^2,
\]
measuring mean squared error between model confidence and binary correctness.

\paragraph{Conflict-aware abstention score.}
The conflict-aware score (CAS) combines raw confidence $c(x)$ with a learned risk signal $r(x) \in [0,1]$:
\[
s_{\mathrm{CAS}}(x) = (1-\alpha)\,c(x) - \alpha\,r(x),
\]
where $r(x)$ is the predicted probability that prediction $x$ is confidently wrong, produced by the within-condition logistic detector, and $\alpha \in [0,1]$ is a fixed mixing weight ($\alpha=0$ recovers confidence-only; $\alpha=1$ yields the detector-only ranker).

\section{Qualitative Case Study Evidence}

\begin{table}[htbp]
\centering
\footnotesize
\setlength{\tabcolsep}{4pt}
\begin{tabular}{p{0.16\linewidth} p{0.78\linewidth}}
\toprule
Item & Verbatim retrieved text \\
\midrule
Question & ``Can exercise lower cholesterol?'' (ground-truth answer: \textsc{YES}) \\
Correct document & ``If your cholesterol numbers aren't where they ought to be, working out should be a key part of your get-healthy strategy. The right kinds of workouts, done regularly, can raise levels of heart-protecting HDL cholesterol and drop dangerous triglyceride levels.'' \\
Incorrect document & ``Exercise itself doesn't directly lower cholesterol but will strengthen the heart and entire circulatory system.'' \\
\bottomrule
\end{tabular}
\caption{Verbatim evidence for the qualitative case study on \textsc{HealthContradict} instance 650. The two retrieved documents directly conflict on whether exercise itself lowers cholesterol.}
\label{tab:qual_case_study_evidence_appendix}
\end{table}

\section{Sensitivity Analyses}

\subsection{Alpha Sensitivity}

\begin{figure}[htbp]
\centering
\includegraphics[width=\textwidth]{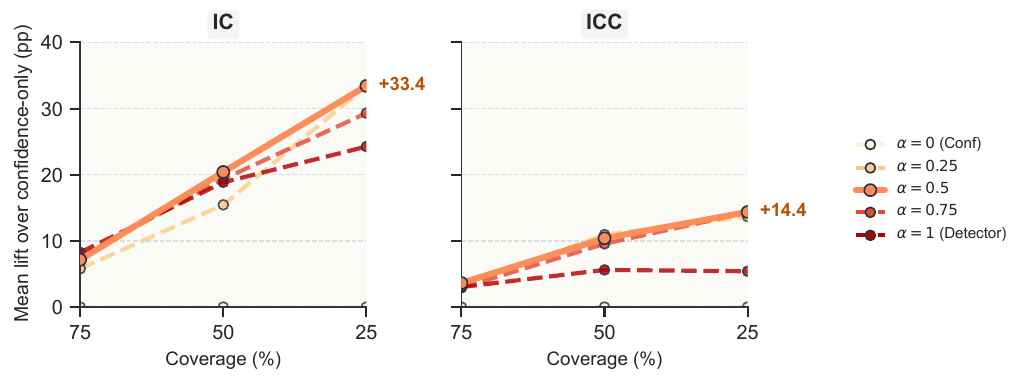}
\caption{Mean held-out selective-accuracy lift of CAS over Conf under the train-threshold transfer protocol as the mixing weight $\alpha$ varies. Curves are shown separately for `IC' and `ICC' at 75\%, 50\%, and 25\% target coverage, with $\alpha=0.5$ visually highlighted as the setting used in the main experiments. The endpoints isolate the two component scores: $\alpha=0$ recovers Conf and so produces a lift of zero by construction, while $\alpha=1$ uses only the conflict detector $r$ and isolates its contribution. Moderate values of $\alpha$ produce similar gains, while $\alpha=1.0$ weakens the more selective `ICC' operating points.}
\label{fig:alpha_sensitivity_selective_gain}
\end{figure}

Figure~\ref{fig:alpha_sensitivity_selective_gain} checks whether the selective-prediction gains depend strongly on the mixing weight $\alpha$, including the two limit cases $\alpha=0$ (Conf alone) and $\alpha=1$ (detector alone). The $\alpha=0$ point sits at zero by construction and anchors the curve at the confidence-only baseline. The $\alpha=1$ point isolates the conflict detector $r$: in `IC', detector-only ranking still gives a mean lift over Conf of 8.3, 18.9, and 24.3 percentage points at 75\%, 50\%, and 25\% coverage, indicating that $r$ carries genuine error-detection signal beyond raw confidence rather than acting as a wrapper around it. In `ICC', $\alpha=1$ retains positive but smaller lift (3.0, 5.6, and 5.4 points), so confidence and the detector are complementary in this condition. Across `IC', all tested $\alpha$ yield substantial positive mean lift; in `ICC', the moderate choices $\alpha \in \{0.25, 0.5, 0.75\}$ give similar gains, whereas $\alpha=1.0$ noticeably weakens the most selective operating point. Together this suggests that using $\alpha=0.5$ in the main experiments is a robust mid-range choice rather than a narrowly tuned optimum, and that the gains over Conf are not solely attributable to recalibration of confidence.

\subsection{Tau Sensitivity}
\label{sec:tau_sensitivity}
\begin{figure}[htbp]
\centering
\includegraphics[width=\textwidth]{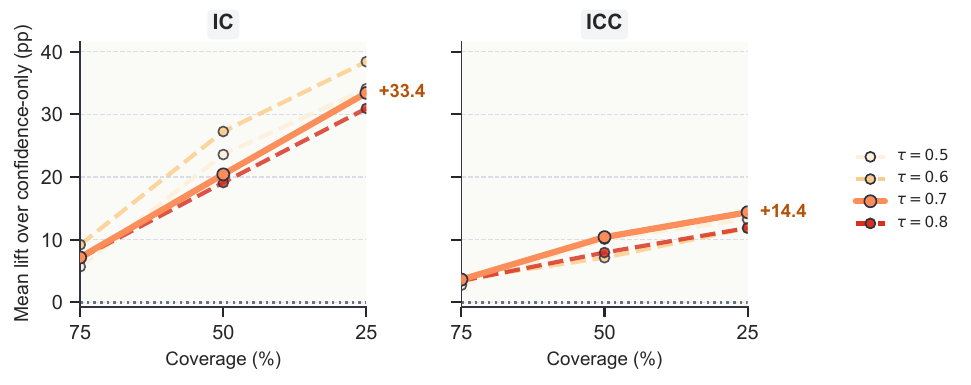}
\caption{Mean held-out selective-accuracy lift of CAS over Conf under the train-threshold transfer protocol as the confidently-wrong supervision threshold $\tau$ varies. Curves are shown separately for `IC' and `ICC' at 75\%, 50\%, and 25\% target coverage, with $\tau=0.7$ visually highlighted as the setting used in the main experiments. Gains remain positive across all tested values, indicating that the method is not tied to a single supervision cutoff.}
\label{fig:conf_threshold_sensitivity_selective_gain}
\end{figure}

Figure~\ref{fig:conf_threshold_sensitivity_selective_gain} checks whether the selective-prediction gains depend strongly on the confidence threshold used to define confidently wrong cases during detector training. In `IC', all tested thresholds yield substantial positive mean lift of CAS over Conf, with $\tau=0.6$ giving the strongest gains in this sweep and $\tau \in \{0.7, 0.8\}$ remaining clearly positive at all three operating points. In `ICC', the pattern is similarly stable: all tested thresholds improve over Conf, with the largest gains at 50\% and 25\% coverage occurring at $\tau=0.7$. These results suggest that the method is reasonably robust to the supervision cutoff, and that using $\tau=0.7$ in the main experiments is a sensible mid-range choice rather than a fragile tuning decision.

\section{Selective Prediction Supplementary Results}

\paragraph{Detector training details.}
The logistic detector is trained separately for each target evidence condition (e.g., `IC' or `ICC') using the model's uncertainty signals (margin, entropy, confidence) together with sentence-embedding features of the question and retrieved document(s). The positive class consists of wrong predictions made with confidence above a threshold $\tau$, with $\tau=0.7$ in the main experiments. Threshold sensitivity is reported in Appendix~\ref{sec:tau_sensitivity}.

As an ablation, we also evaluate a cross-condition predictor that uses the same target-condition features but replaces direct labels with a proxy label based on confidence drop relative to the corresponding `CC' prompt: for an instance $x$, the proxy label is 1 when $c_{\text{CC}}(x) - c_{\text{target}}(x) > 0.3$, asking whether a large drop from correct-only evidence is sufficient to identify risky cases.

We use an 80/20 train/test split. On the training split, we obtain 5-fold out-of-fold scores and choose thresholds for target coverages $\{0.75, 0.5, 0.25\}$, meaning that the system answers the top 75\%, 50\%, or 25\% of ranked examples and abstains on the rest. These thresholds are transferred unchanged to the held-out test split, so the realized answered fraction on test need not equal the target exactly; Appendix Table~\ref{tab:selective_prediction_appendix} therefore reports realized coverage for both CAS and Conf.

\paragraph{Results.}
This section reports the same selective-prediction protocol used in the main text. For each target coverage, we choose a CAS threshold from train out-of-fold predictions and then apply that same numeric cutoff to the held-out test split. The test set therefore does not have to land at exactly the nominal 75\%, 50\%, or 25\% coverage: the score distribution can shift from train to test, and score ties near the cutoff can move a few examples across the boundary. We evaluate Conf at approximately matched test coverage so that the comparison is made at nearly the same operating point. Reporting realized coverage is important because otherwise a small accuracy difference could simply reflect one method answering more examples than the other. Table~\ref{tab:selective_prediction_appendix} therefore reports both selective accuracy and realized coverage for CAS and Conf.

\begin{table}[t]
\centering
\small
\setlength{\tabcolsep}{3.5pt}
\resizebox{\textwidth}{!}{%
\begin{tabular}{lcccccccccccc}
\toprule
& \multicolumn{6}{c}{`IC'} & \multicolumn{6}{c}{`ICC'} \\
\cmidrule(lr){2-7}\cmidrule(lr){8-13}
& \multicolumn{2}{c}{75\%} & \multicolumn{2}{c}{50\%} & \multicolumn{2}{c}{25\%} & \multicolumn{2}{c}{75\%} & \multicolumn{2}{c}{50\%} & \multicolumn{2}{c}{25\%} \\
\cmidrule(lr){2-3}\cmidrule(lr){4-5}\cmidrule(lr){6-7}\cmidrule(lr){8-9}\cmidrule(lr){10-11}\cmidrule(lr){12-13}
Model & Comb. & Conf. & Comb. & Conf. & Comb. & Conf. & Comb. & Conf. & Comb. & Conf. & Comb. & Conf. \\
\midrule
Llama-3.1-8B & 0.377 / 0.793 & 0.315 / 0.793 & 0.485 / 0.538 & 0.303 / 0.538 & 0.636 / 0.239 & 0.304 / 0.250 & 0.703 / 0.750 & 0.645 / 0.750 & 0.772 / 0.500 & 0.677 / 0.505 & 0.891 / 0.299 & 0.702 / 0.310 \\
Meditron3-8B & 0.421 / 0.788 & 0.386 / 0.788 & 0.538 / 0.505 & 0.366 / 0.505 & 0.619 / 0.228 & 0.262 / 0.228 & 0.780 / 0.766 & 0.752 / 0.766 & 0.878 / 0.489 & 0.761 / 0.500 & 0.976 / 0.228 & 0.791 / 0.234 \\
Phi-4 & 0.489 / 0.723 & 0.451 / 0.723 & 0.615 / 0.424 & 0.418 / 0.429 & 0.839 / 0.168 & 0.339 / 0.337 & 0.769 / 0.728 & 0.687 / 0.728 & 0.851 / 0.473 & 0.667 / 0.473 & 0.848 / 0.250 & 0.717 / 0.250 \\
Qwen3-4B-Inst. & 0.451 / 0.783 & 0.347 / 0.783 & 0.595 / 0.457 & 0.347 / 0.783 & 0.667 / 0.261 & 0.347 / 0.783 & 0.776 / 0.799 & 0.764 / 0.804 & 0.877 / 0.440 & 0.772 / 0.739 & 0.976 / 0.223 & 0.772 / 0.739 \\
Qwen3-8B & 0.632 / 0.783 & 0.556 / 0.783 & 0.753 / 0.527 & 0.594 / 0.549 & 0.805 / 0.223 & 0.598 / 0.473 & 0.867 / 0.777 & 0.819 / 0.783 & 0.960 / 0.549 & 0.851 / 0.658 & 0.980 / 0.277 & 0.851 / 0.658 \\
Qwen3.5-9B & 0.464 / 0.679 & 0.349 / 0.701 & 0.662 / 0.418 & 0.395 / 0.440 & 0.789 / 0.207 & 0.500 / 0.239 & 0.879 / 0.717 & 0.889 / 0.734 & 0.944 / 0.484 & 0.929 / 0.533 & 0.976 / 0.223 & 0.951 / 0.223 \\
\midrule
Mean & 0.472 / 0.758 & 0.401 / 0.762 & 0.608 / 0.478 & 0.404 / 0.541 & 0.726 / 0.221 & 0.392 / 0.385 & 0.796 / 0.756 & 0.759 / 0.761 & 0.880 / 0.489 & 0.776 / 0.568 & 0.941 / 0.250 & 0.797 / 0.402 \\
\bottomrule
\end{tabular}
}
\caption{Held-out selective prediction results underlying Figure~\ref{fig:selective_prediction_curves}. Each cell reports selective accuracy / realized coverage for CAS and Conf. CAS thresholds are selected on train out-of-fold scores and transferred unchanged to test; Conf is evaluated at approximately matched test coverage, so realized coverages may differ slightly.}
\label{tab:selective_prediction_appendix}
\end{table}

\end{document}